\documentclass[10pt,twocolumn,letterpaper]{article}

\usepackage{iccv}
\usepackage{times}
\usepackage{epsfig}
\usepackage{graphicx}
\usepackage{amsmath}
\usepackage{amssymb}
\usepackage{color}
\usepackage[labelsep=period]{caption}
\usepackage{subfigure}
\usepackage{caption}
\usepackage{cite}
\usepackage[table]{xcolor}
\usepackage{multirow}

\usepackage{url}
\usepackage[toc,page]{appendix}
\usepackage{pifont}
\usepackage[pagebackref=true,breaklinks=true,letterpaper=true,colorlinks,bookmarks=false]{hyperref}
\usepackage{array}
\usepackage{enumitem}

\newcommand{\Cgray}[1]{\cellcolor[HTML]{D8D6D6}} 

\newcommand{\chapternote}[1]{{%
		\let\thempfn\relax
		\footnotetext[0]{{#1}}
	}}

\def\iccvPaperID{5} 

\iccvfinalcopy

\begin{document}
	
	\title{Unified Framework for Automated Person Re-identification and \\ Camera Network Topology Inference in Camera Networks}
	
	\author{\qquad Yeong-Jun Cho, Jae-Han Park*, Su-A Kim*, Kyuewang Lee and Kuk-Jin Yoon\\
		Computer Vision Laboratory, GIST, South Korea \\
		{\tt\small\qquad $\lbrace${yjcho, qkrwogks, suakim, kyuewang, kjyoon}$\rbrace$@gist.ac.kr}\\
		\and
	}

	\maketitle

	\begin{abstract}
		
		Person re-identification in large-scale multi-camera networks is a challenging task because of the spatio-temporal uncertainty and high complexity due to large numbers of cameras and people. To handle these difficulties, additional information such as camera network topology should be provided, which is also difficult to automatically estimate. 
		In this paper, we propose a unified framework which jointly solves both person re-id and camera network topology inference problems. 
		The proposed framework takes general multi-camera network environments into account.  
		To effectively show the superiority of the proposed framework, 
		we also provide a new person re-id dataset with full annotations, named {\small\texttt{SLP}}, captured in the synchronized multi-camera network.
		Experimental results show that the proposed methods are promising for both person re-id and camera topology inference tasks.
		
	\end{abstract}
	\vspace{-0pt}
	
	\pagestyle{empty}
	
	\section{Introduction}
\begin{figure}[t]
		\centering
		\includegraphics[width=1\columnwidth]{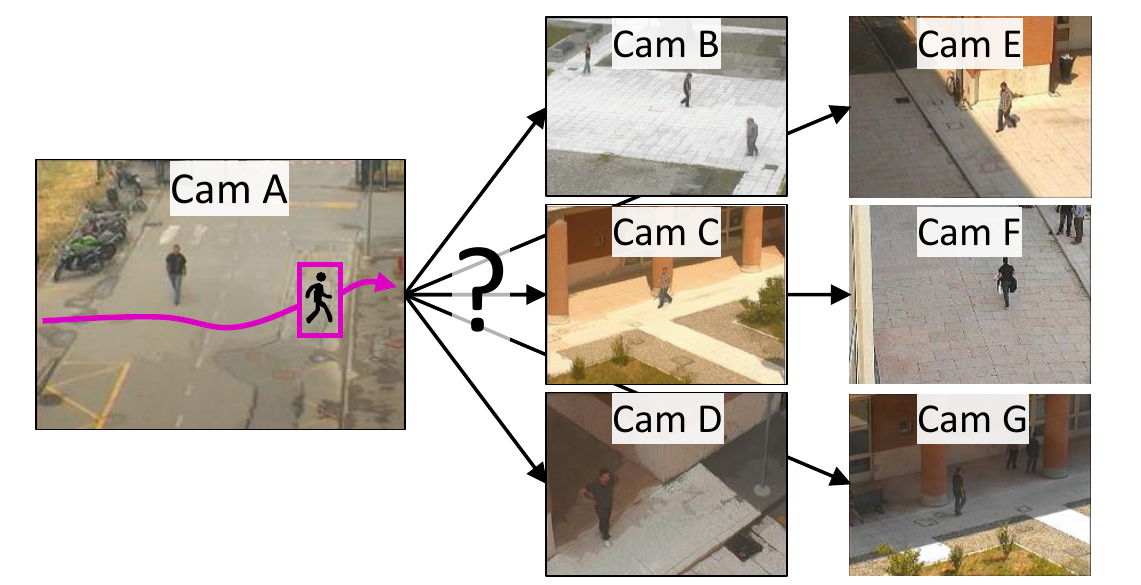}
		\caption{Challenges of large-scale person re-identification : spatio-temporal uncertainties between cameras.}
		\label{fig_1}
		\vspace{-0pt}
	\end{figure}

	\thispagestyle{empty}
	\chapternote{* These authors contributed equally to this work. This work has been done while Su-A Kim and Kyuewang Lee were with Computer Vision Lab in GIST.
		Currently Su-A Kim is with Intel VCI (Saarland Informatics Campus) and supported by the European Commission through the H2020-MSCA Distributed 3D Object Design (Grant No. 642841) since April 2017. Currently Kyuewang Lee is with ASRI in seoul national university. Kuk-Jin Yoon is the corresponding author.}
	Person re-identification (\textit{i.e.} re-id) is the task of \textit{automatically} recognizing and identifying a person across multiple views in multi-camera networks, and has been studied for last decades. 
	Nevertheless, the re-id in large-scale multi-camera networks still remains a challenging task because of the large spatio-temporal ambiguity and high complexity due to large numbers of cameras and people. Especially, it becomes more challenging when camera views are not overlapped each other.
	As shown in Fig.~\ref{fig_1}, spatio-temporal uncertainty due to the unknown geometrical  relationship between cameras in a multi-camera network makes the re-id difficult. Unless some prior knowledge about the camera network is given, re-id should be done by thoroughly matching the person of interest with all other people appeared in the other camera within some time interval.
	This exhaustive search method is slow in general and shows unsatisfactory results in many cases because it is hard to find a correct match among a large number of candidates --- among the large number of candidates, there might be many people having similar appearances with the person of interest.
	
	
	However, most of the previous works~\cite{farenzena2010person, davis2007information} conduct the exhaustive search to re-identify people across multiple cameras, relying solely on the appearance information of people.
	These methods work quite well when the numbers of cameras and people are small, but cannot effectively handle the aforementioned challenges in the large-scale problem.
	To resolve the complexity problem and to improve the accuracy of the large-scale person re-id, the number of matching candidates of a person of interest should be constrained and reduced by inferring and exploiting the spatio-temporal relation between cameras, referred to as the camera network topology.	
	In recent years, several camera network topology inference methods~\cite{makris2004bridging, loy2010time} have been proposed.
	Those methods infer the topology of a camera network based on the simple occurrence correlation between entering and exiting events of people.    
	However, since they do not perform any appearance-based validation for topology inference, the inferred topology can be inaccurate in crowded scenes. 
	
	The main idea of this paper is that the camera network topology inference and person re-id can be solved jointly while complementing each other. 
	Based on this idea, we propose a unified framework which automatically solves both person re-id and camera network topology inference problems together.
	To the best of our knowledge, this is the first attempt to solve both problems jointly.
	In the proposed framework, we first infer the initial camera network topology using only highly reliable re-id results obtained by the proposed multi-shot matching method. This initial topology is used to improve the person re-id results, and the improved re-id results are then used to refine the camera network topology. This procedure is repeated until the estimated camera network topology converges. Once we estimate the reliable camera network topology in the training stage, we can utilize it for the online person re-id and update the camera network topology with time. 
	
	To sum up, we propose a multi-shot person re-id method which exploits time-efficient random forest (Sec.~\ref{subsec:RF_person_reid}).
	We also propose fast and accurate camera network topology inference method in Sec.~\ref{subsec:topology_infer}.
	It is worthy to note that our proposed framework runs fully automatic with minimal prior knowledge about the environments.
	Besides the proposed methods, we also provide a new synchronized large-scale person re-id dataset named {\texttt{SLP}} (Sec.~\ref{sec:Pe-Lar_database}). 
	To validate our unified framework, we extensively evaluate the performance of the proposed method and compare with other state-of-the-art methods.  
	
	\section{Previous Works}
	\label{sec:preivous}
	
	Person re-id methods can be categorized into non-contextual and contextual methods as summarized in~\cite{bedagkar2014survey}.
	In general, non-contextual methods rely only on appearances of people and measure visual similarities between people to establish correspondences, while contextual methods exploit additional contexts such as human pose prior, camera parameters, geometry, and camera topology.

	\noindent \textbf{Non-contextual Methods} \quad
	In order to identify people across non-overlapping views, most of works generally rely on appearances of people by utilizing appearance-based matching methods with feature learning or metric learning.	
	For the feature learning, many works~\cite{farenzena2010person, liu2012person, zhao2014learning} have tried to design visual descriptors to well describe the appearance of people.
	Regarding the metric learning, several methods  such as KISSME~\cite{koestinger2012large} and LMNN-R~\cite{dikmen2011pedestrian} have been proposed and applied to the re-id problem~\cite{weinberger2005distance, roth2014mahalanobis}. 
	
	Although many non-contextual methods have improved the performance of person re-id, the challenges such as spatio-temporal uncertainty between non-overlapping cameras and high computational complexity still remain.

	\noindent \textbf{Contextual Methods} \quad   
	Several works~\cite{bak2015person, cho2016improving, wu2015viewpoint} using human pose priors have been proposed recently.
	Exploiting human poses mitigates ambiguities by seeking pose variations of people but spatio-temporal ambiguities between cameras still remain. 
	
	To resolve the spatio-temporal ambiguities, many works have tried to employ camera network topology and camera geometry. 
	Several works~\cite{javed2003tracking,cai2014exploring,rahimi2004simultaneous} assume that the camera network topology is given, and show the effectiveness of the topological information.
	However, the topology is not given in the real-world scenario; thus, many works have tried to infer the camera network topology in an unsupervised manner.
	Makris~\textit{et al.}~\cite{makris2004bridging} proposed a topology inference method that simply observes entering and exiting events of targets and measures correlations between the events to establish the camera network topology. This method was extended in~\cite{niu2006recovering,stauffer2005learning,chen2014object}. 
	Similarly, Loy~\textit{et al.}~\cite{loy2010time} also proposed topology inference methods to understand multi-camera activity by measuring correlation or mutual information between simple activity patterns.
	
	The aforementioned topology inference methods~\cite{makris2004bridging, stauffer2005learning,chen2014object, loy2010time}, so-called, event-based approaches, are practical since they do not require any appearance matching steps such as re-id or inter-camera tracking for topology inference.
	However, the topology inferred by the event-based approach can be inaccurate since the topology may be inferred from false event correlations.

	\section{Proposed Unified Framework}
	\label{sec:proposed}
\begin{figure}[t]
		\centering
		\vspace{-0pt}        
		\includegraphics[width=1\columnwidth]{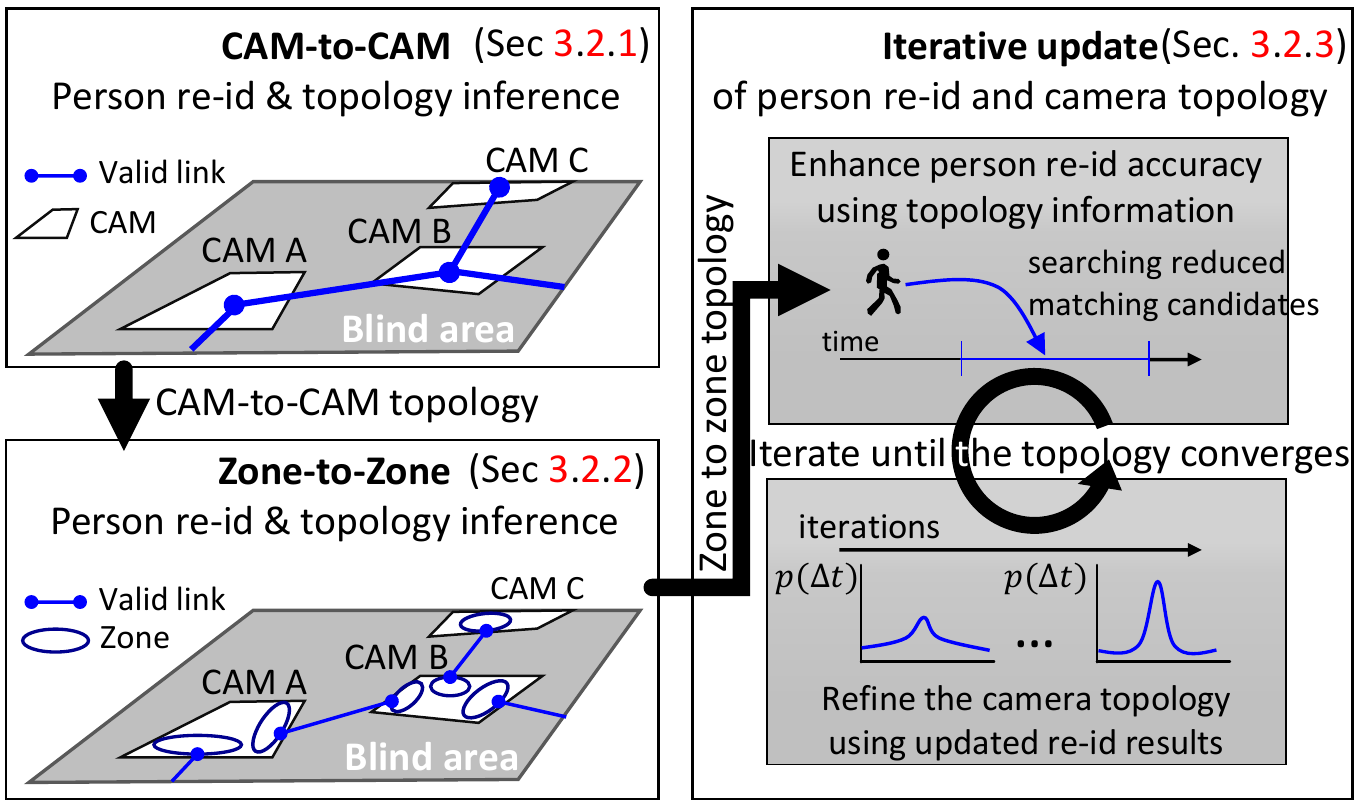} 		\vspace{-0pt}
		\vspace{-0pt}  
		\caption{The proposed unified framework for person re-identification and camera network topology inference.}\vspace{-0pt}
		\label{fig_2}
		\vspace{-0pt}  
	\end{figure}

	Figure~\ref{fig_2} illustrates the proposed unified framework for person re-id and camera network topology inference.
	In the proposed framework, we first train random forest-based person classifiers~(Sec.~\ref{subsec:RF_person_reid}) for efficient person re-id. Subsequently, we jointly estimate and refine the camera network topology and person re-id results~(Sec.~\ref{subsec:topology_infer}) using the trained random forests.

	\subsection{Random Forest-based Person Re-identification}
	\label{subsec:RF_person_reid}
	Most of previous works mainly focused on enhancing re-id performance. However, when handling a large number of people, time complexity is also very important for building a practical re-id framework. 
	To this end, we utilize a random forest algorithm~\cite{breiman2001random} for the multi-shot person re-id and incorporate it into our framework.
	We denote the $k$-th appearance of person $i$ in camera $c_{A}$ as ${\mathbf{ v }}^{ c_{A} }_{i,k}$. A set of the appearances of people in the camera is expressed as,
	\vspace{-0pt}
	\begin{equation}
		\label{eq:1}
		\mathcal{D}^{c_{A}}={ \left\{ { \left( {\mathbf{ v }}^{ c_{A} }_{i,k},{ y }_{ i } \right)  } | { 1 \leq i \leq N^{c_{A}}, 1 \leq k \leq K^{c_{A}}_i}\right\}},
	\end{equation}
	where $y_{i}$ is the label of person $i$ is the number of people in camera $c_{A}$ and $K^{c_{A}}_i$ is the number of appearances of person $i$ in camera $c_{A}$. We then train a random forest classifier using the appearance set $\mathcal{D}^{c_{A}}$.
	After the random forest classifier is trained, we have the probability distribution of classification as $p^{c_{A}}\left( { y }|{ \mathbf{v} } \right)$.
	
	To obtain a multi-shot person re-id result, we test multiple appearances of each person and average the multiple results as $p^{{c}_{A}}( y |{ {\mathbf{v}}^{{c}_{B}}_{j} } ) =\frac { 1 }{ K_{j}^{c_{B}} }\sum _{ l=1 }^{ K_{j}^{c_{B}} }{ p^{{c}_{A}}( y |{ {\mathbf{v}}^{{c}_{B}}_{ j,l } } )  },$
	where $K_{j}^{c_{B}}$ is the number of appearances of the probe.
	Among the probability distribution $p^{{c}_{A}}\left( y |{ {\mathbf{v}}^{{c}_{B}}_{j} }\right)$, 
	we choose a final matched label ${y}_{i}^{*}$ which maximizes $p^{{c}_{A}}( y_{i} |{ {\mathbf{v}}^{{c}_{B}}_{j} } )$.
	As the result of the multiple appearance matching test, we have a corresponding pair~($\mathbf{ v }_{{y}_{i}^{*}}^{{c}_{A}}$, $\mathbf{ v }_{j}^{ { c }_{ B } }$) between camera $c_{A}$ and $c_{B}$. 
	Finally, we calculate a similarity score of the corresponding pair by selecting the smallest matching score as in~\cite{farenzena2010person}. 
	We denote the similarity score as $S\left( \mathbf{ v }_{ {y}_{i}^{*} }^{ { c }_{ A } }, \mathbf{ v }_{ j }^{ { c }_{ B } } \right)$. The score lies on $[0,1]$ and it is used in Sec.~\ref{subsec:topology_infer} for inferring the topology.
	The tree structure of the random forest method makes the multi-shot test very fast.
	Besides the superiority of the computational cost, our method gives high person re-id accuracy as shown in Sec.~\ref{subsec:exp:cam_to_cam}.	
	
	\subsection{Camera Network Topology Initialization}
	\label{subsec:topology_infer}

	{Camera network topology} represents spatio-temporal relations and connections between cameras in the network.
	It involves with the inter-camera transition distributions between two cameras, which denote transition distributions of objects across cameras according to time, and represents the strength of connectivity between cameras.
	In general, the topology is represented as a graph $G=\left(V, E\right)$, where vertices $V$ denote cameras and edges $E$ denote inter-camera transition distributions as shown in Fig.~\ref{fig_10} (b).

	\subsubsection{CAM-to-CAM topology inference}
	\label{subsubsec:CAMCAM_con_analy}

	First, we estimate transition distributions between cameras to build the CAM-to-CAM topology. 
	In this work, we estimate the transition distributions based on the person re-id results.
	We first split a whole group of people into several sub-groups according to their time-stamps and train a series of random forest classifiers with time window $T$.
	Next, we search correspondences of people disappeared in a camera using the trained random forest classifiers of other cameras. 
	Initially, we have no transition distributions between cameras to utilize. 
	Hence, we consider every pair of cameras in the camera network and search correspondences within wide time interval. 
	When a person disappears at time $t$ in a certain camera, we search the correspondence of the person from the other cameras within time range $[t-T,t+T]$. 

\begin{figure}[t]
		\centering
		\subfigure[CAM1 -- CAM2]{\includegraphics[width=0.49\columnwidth]{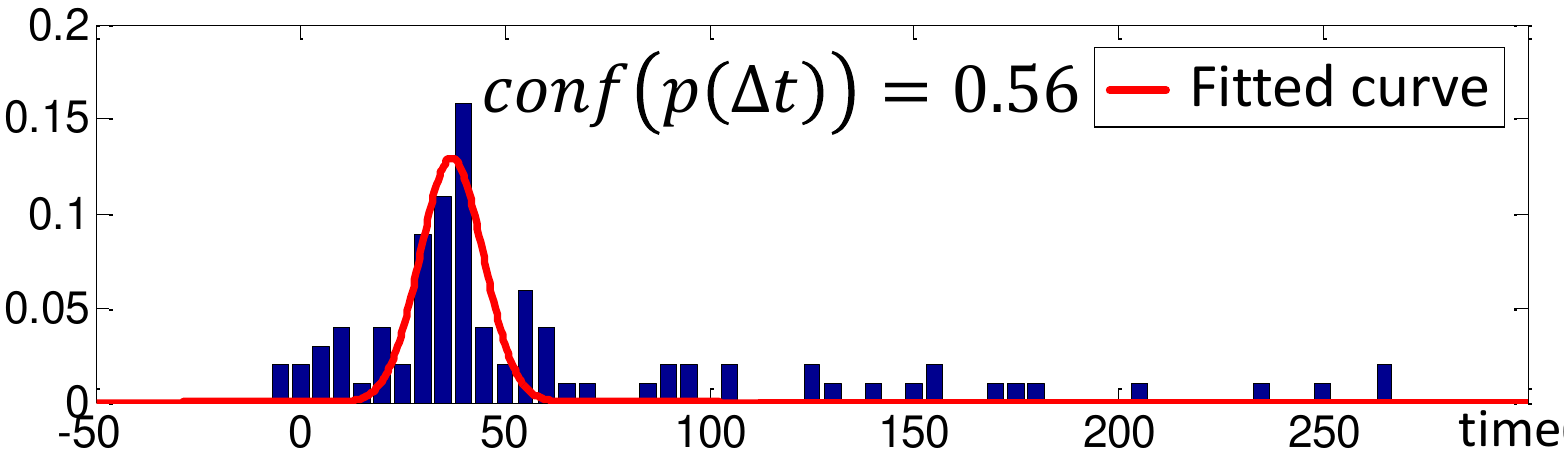}}
		\subfigure[CAM1 -- CAM4]{\includegraphics[width=0.49\columnwidth]{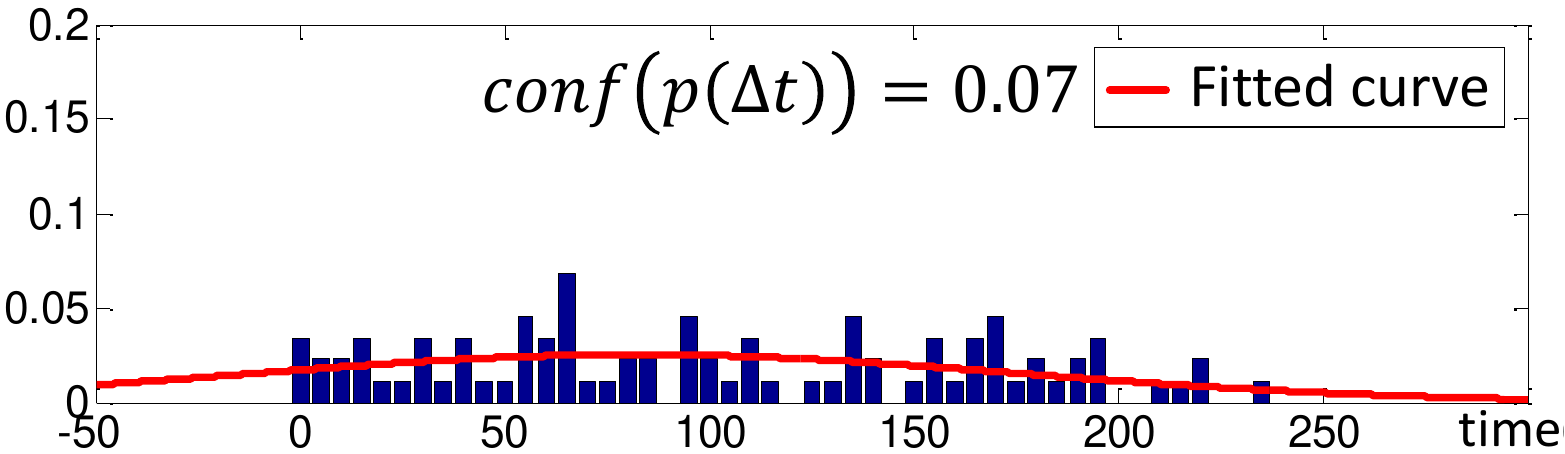}}
		\vspace{-0pt}
		\caption{Examples of estimated transition distributions with connectivity confidences.}
		\vspace{-0pt}
		\label{fig_5}
	\end{figure}

	When these initial correspondences are given, we infer transition distributions between cameras using highly reliable correspondences only.
	We regard a correspondence as a reliable one with a high similarity score when $S( \mathbf{ v }_{ { y }_{i}^{ * } }^{ { c }_{ A } }, \mathbf{ v }_{ j }^{ { c }_{ B } })> \theta_{sim}$.
	Transition distribution inference procedure is as follows:
	(1) calculating time difference between correspondences and making a histogram of the time difference; 
	(2) normalizing the histogram by the total number of reliable correspondences. 
	We denote the transition distribution as $p\left(\Delta t\right)$.
	Figure~\ref{fig_5} shows two distributions: Fig.~\ref{fig_5}~(a) comes from a pair of cameras having strong connection, and Fig.~\ref{fig_5}~(b) from a pair of cameras having a weak or no connection.

	\noindent \textbf{Connectivity Check} \quad Based on the estimated transition distributions, we automatically identify whether a pair of cameras is connected or not. We assume that the transition distribution follows a normal distribution if there is a topological connection.
	Based on this assumption, we fit a Gaussian model $N(\mu,\sigma^2)$ to the distribution $p\left(\Delta t\right)$ and measure the connectivity of a pair of cameras based on the following observations:

	\noindent $\bullet$ \textbf{Variance of $p\left(\Delta t\right)$}: In general, most of  people re-appear around the certain transition time $\mu$; therefore, the variance of the transition distribution $(\sigma^2)$ is not extremely large and the distribution shows a clear peak.  
	
	\noindent $\bullet$ \textbf{Fitting error}: Although the distribution comes from a pair cameras with a weak connection, the variance of the distribution can be small and the distribution can have a clear peak due to noise. 
	In order to measure the connectivity robust to noise, we consider the model fitting error $\mathcal{E}\left( p\left(\Delta t\right) \right)\in[0,1]$ calculated by R-squared statistics.   
	
	Based on the above observations, we newly define a connectivity confidence between a pair of cameras as
	\begin{equation}
		conf\left( p\left(\Delta t\right) \right) = e^{-\sigma}\cdot \left(1-\mathcal{E}\left( p\left(\Delta t\right) \right)\right).
	\end{equation}
	The connectivity confidence lies on $[0, 1]$. 
	We regard a pair of cameras as a valid link when $conf\left(p\left(\Delta t\right)\right)>\theta_{conf}$.
	Compared with a previous method~\cite{makris2004bridging} which only considers the variance of a distribution, our method is more robust to noise of distributions.
	Using the defined confidences, we check every pair of cameras and reject invalid links as shown in Fig.~\ref{fig_10}. We can see that many camera pairs in the camera network have weak connection; therefore, we can greatly reduce computation time and save resources. Only the valid pairs of cameras are proceeding to the next step.
	
	\vspace{-0pt}
	
	\vspace{0pt}			
	\subsubsection{Zone-to-Zone topology inference} \label{subsubsec:ZonZon_con_analy}
	
	In this step, we estimate transition distributions between zones in cameras and build a Zone-to-Zone topology.
	For each camera, a set of entry and exit zones is automatically learned by \cite{makris2003automatic}.
	Note that, we only consider exit-to-entry zone pairs when two zones belong to different cameras. 
	Other pairs of zones such as exit-to-exit, entry-to-entry, and entry-to-exit are not considered.
	
	A person disappeared at an exit zone at time $t$ is likely to appear at entry zone in a different camera within the certain time interval $T$. Therefore, we search the correspondence of the disappeared person from entry zones in different cameras within time range $[t,t+T]$.
	Similarly as in Sec.~\ref{subsubsec:CAMCAM_con_analy}, we train a series of random forest classifiers for each entry zone and measure connectivity confidences of all possible pairs of zones using only reliable correspondences.
	Through this step, many invalid pairs of zones between cameras are ignored.
	In the next section, we iteratively update the valid links between zones and build a camera topology map of the camera network.

	\subsubsection{Iterative update of person re-identification \\ and camera network topology} 
	\label{subsubsec:iter_topology_infer}
	
	After the Zone-to-Zone topology inference, we have an initial topology map between every pair of zones in the camera network.
	However, the initial topology map can be inaccurate, since it is inferred by noisy initial re-id results.
	As mentioned before, camera network topology information and re-id results can be used for each other: the inferred topology of the camera network can enhance the person re-id performance, and person re-id can assist the topology inference. Therefore, we update person re-id results and the camera network topology in an iterative manner as follows:


		\begin{table*}[t]
			\centering
			\caption{Details of our new dataset: \texttt{SLP}.} \vspace{-0pt}
			\label{tab_1}
			\footnotesize{
				\renewcommand{\arraystretch}{0.9} 
				\begin{tabular}{r||c|c|c|c|c|c|c|c|c|c}
					\noalign{\hrule height 1pt}
					Index        & CAM 1         & CAM 2           & CAM 3            & CAM 4           & CAM 5           & CAM 6           & CAM 7          & CAM 8        & CAM 9    & Total \\ \hline\hline
					\# ID        &  256          & 661             & 1,175            & 243             & 817             & 324             & 516            & 711          & 641      &  2,632 \\ 
					\# frames    &  19,545       & 65,518          & 104,639          & 41,824          & 78,917          & 79,974          & 93,978         & 53,621       & 42,347   &  580,363 \\ 
					\# annotated box & 47,870    & 205,003         & 310,262          & 65,732          & 307,156         & 160,367         & 78,259         & 176,406      & 117,087  &  1,468,142 \\ 
					Duration     & 2h 13m        & 2h 12m              & 2h 22m              & 2h              & 2h 21m              & 2h              & 2h 38m              & 2h 29m             & 2h 28m     &  --  \\ \noalign{\hrule height 1pt} 
				\end{tabular}
			}
			\vspace{-0pt}
		\end{table*}

	\begin{itemize}[leftmargin=10pt]
		\setlength\itemsep{-0.3em}
		\item \textbf{Step 1.} Update the time window $T$. 
		The initial time window $T$ was set quite wide, but now we can narrow it down based on the inferred topology. 
		To this end, we find lower and upper time bounds $\left(T_{L},T_{U}\right)$ of the transition distribution $p\left(\Delta t\right)$ with  a constant $R$ as,
		\begin{equation}
			p\left( T_{L}\le \Delta t\le T_{U} \right) = { \frac{R}{100} }.
		\end{equation}
		We set $R$ as $95$, following 3-sigma rule, in order to cover the most of the distribution (95\%) and ignore some outliers (5\%). Then, using the obtained time bounds, the time window $T$ is updated as,
		\begin{equation}
			\label{eq:6}
			T = \frac { 1 }{ 1-\mathcal{E}\left( p\left(\Delta t\right) \right) } \left(T_{U}-T_{L} \right),
		\end{equation}
		where $\mathcal{E}\left( p\left(\Delta t\right) \right)$ is a topology fitting error rate. When the fitting error is large, the time window $T$ becomes large. Thanks to our update strategy, we can avoid the overfitting of the topology during the update steps.
		\item \textbf{Step 2.} Re-train a series of random forests of an entry zone with the updated time window $T$. 
		\item \textbf{Step 3.} Find correspondences of disappeared people at an exit zone. Based on the topology, a person disappeared at time $t$ at the exit zone is expected to appear around the time $(t+\mu)$ at the entry zone of the other camera. 
		Using the topological information, we search the correspondence of the person from a trained random forest having the center of time slot close to $(t+\mu)$.
		\item \textbf{Step 4.} Update a topology using reliable correspondences with a high similarity score $S( \mathbf{ v }_{ { y }_{i}^{ * } }^{ { c }_{ A } }, \mathbf{ v }_{ j }^{ { c }_{ B } } )>\theta_{sim}$.
	\end{itemize}
	
	\vspace{-0pt}
	
	This procedure~(Step 1 -- Step 4) is repeated until the topology converges. The above procedure improves the performance of re-id as well as the accuracy of the topology inference.
	We empirically set parameters $\theta_{sim}$, $\theta_{conf}$, $T$ as $0.7$, $0.4$, $600$ based on an extensive evaluation. 

	\section{A New Person Re-id Dataset : \texttt{SLP}}	
	\label{sec:Pe-Lar_database}

	To validate the performance of person re-id methods, numerous datasets have been published. 
	For example, \cite{gray2007evaluating} was constructed by two cameras and contains 632 people. Each camera provides one single image for one person. 
	On the contrary, \cite{bialkowski2012database} includes 150 people captured from eight cameras. In this dataset, each camera provides multiple images for a person.
	However, despite the outburst of published datasets, none of them reflect practical large-scale surveillance scenarios, in which (1) video frames captured from multiple synchronized cameras are available, and (2) both the numbers of people and cameras are large.

	Most of the public datasets include a small number of people~(\textit{\# IDs} $<200$)~\cite{Zheng2009associating, cheng2011custom, baltieri2011_308,  bialkowski2012database} or cameras~(\textit{\# cam} $<5$)~\cite{gray2007evaluating, Zheng2009associating, cheng2011custom, li2014deepreid, Chen2015An, wang2014person}. Moreover, some datasets provide single-shot of each person~\cite{gray2007evaluating, loy2010time} or do not provide annotation information of people (\textit{track gt}) throughout the entire video sequences~\cite{Zheng2009associating, cheng2011custom, baltieri2011_308, li2014deepreid, zheng2016person}. Furthermore, there are only a few datasets which provide camera synchronization information or time stamps of all frames~(\textit{sync})~\cite{Chen2015An}. 
	
\begin{figure}[t]
	\centering
	\subfigure[Layout of a camera network]{\includegraphics[width=0.75\columnwidth]{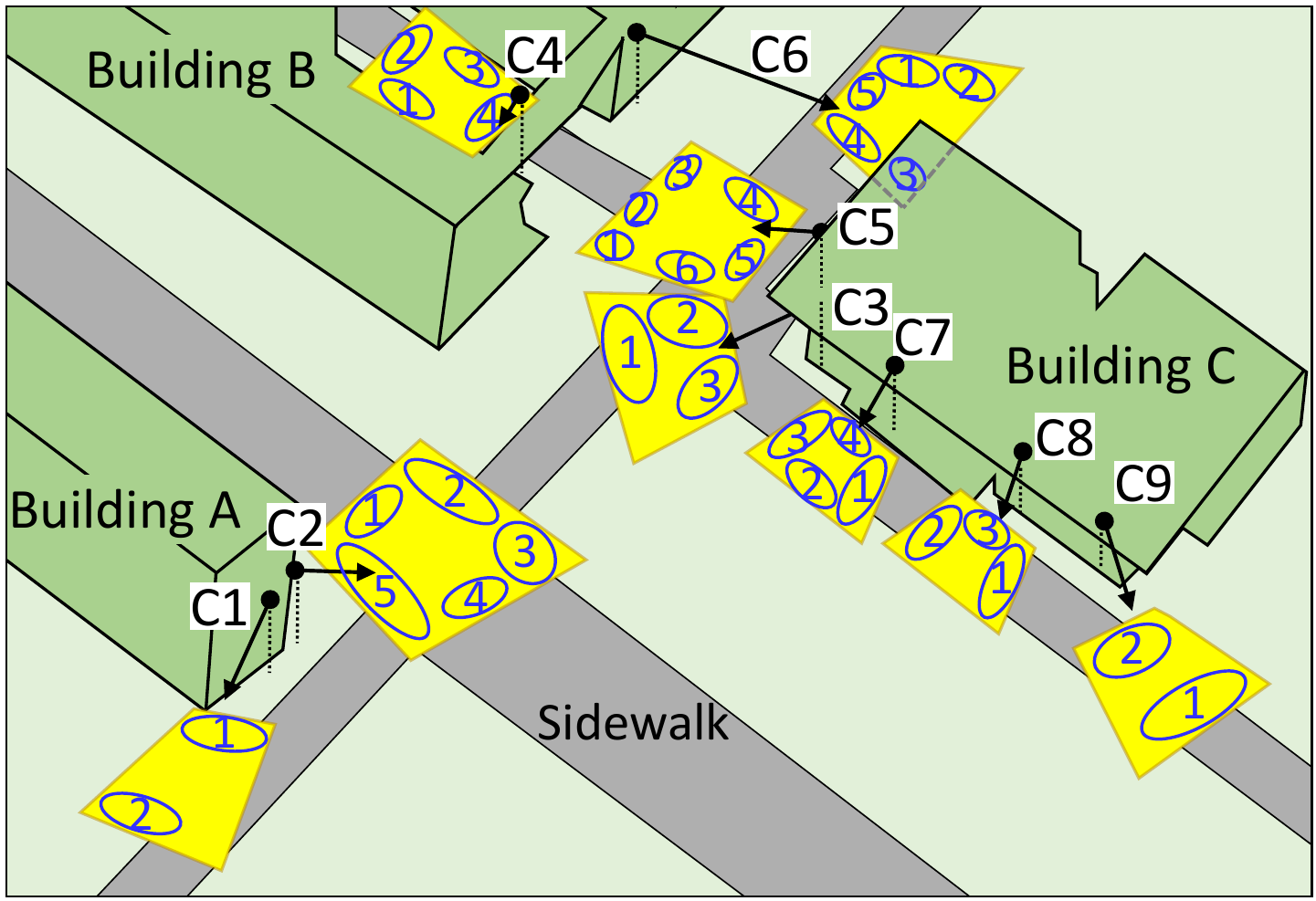}} \vspace{-0pt}
	\subfigure[Example frames of nine cameras]{\includegraphics[width=0.75\columnwidth]{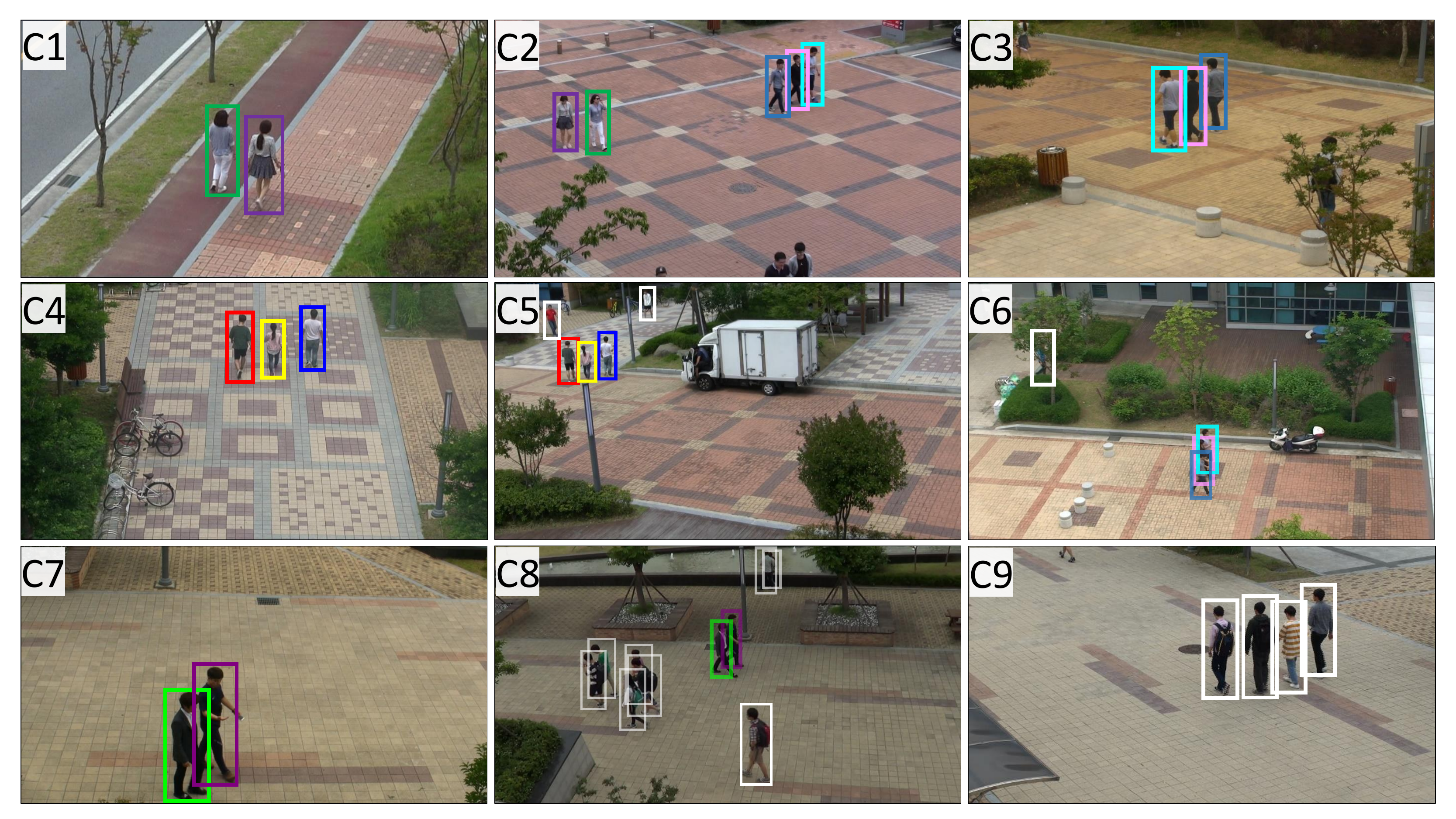}}
	
	\caption{A new synchronized large-scale person re-identification dataset: \texttt{SLP}.}
	\label{fig_8} \vspace{-0pt}
\end{figure}

	In this paper, we provide a new synchronized large-scale person re-id dataset called \texttt{SLP} constructed for practical large-scale surveillance scenarios. The main characteristics of our dataset are as follows:
	The total number of people in the dataset is 2,632. The layout of the camera network and example frames are shown in Fig.~\ref{fig_8}. It provides extracted feature descriptor of each person\footnote{In this version, we do not provide the entire video frames but provide extracted feature descriptors due to legal problem. However we will provide entire video frames in the near future.}. The ground truth of every person is available. Table~\ref{tab_1} shows the details of \texttt{SLP}. It is available on online. {\small\url{https://sites.google.com/view/yjcho/project-pages/re-id_topology}}.

			\section{Experimental Results}
			\label{sec:exp}

			\noindent \textbf{Experimental settings}
			
			Since we mainly focus on person re-identification and camera topology inference problems, we assume that person detection and tracking results are given.
			We divide our dataset into two subsets according to time: The first subset contains 1-hour data starting from the global start time~(AM 11:20). It is used in an camera network topology training stage. 
			The latter subset including the remaining data is utilized in a person re-id test stage.  
			We used LOMO feature~\cite{liao2015person} to describe the appearances of people. Note that our method can adopt any kind of feature extraction methods.
			
			\noindent \textbf{Evaluation methodology}
			
			To evaluate the performance of person re-id, we measure the re-id accuracy (Re-id acc) defined as $\frac {TP}{T_{gt}}$, where $TP$ is the number of true matching results and $T_{gt}$ is the total number of ground truth re-id pairs in the camera network.
			To evaluate the accuracy of the camera network topology, we measure topology distance (Top dist).
			When an inferred transition distribution and a ground truth are given as $(p(\Delta t)$$\sim$$N\left(\mu, \sigma^{2}\right),p_{gt}(\Delta t)$$\sim$$N\left(\mu_{gt}, \sigma_{gt}^{2}\right))$, we defined the topology distance based on \textit{Bhattacharyya} distance which measures the difference between two probability distributions as $d_B(p,p_{gt}){=}-\ln \left(\int \sqrt{p(\Delta t)p_{gt}(\Delta t)}\,\text{d}\Delta t \right)$. If there are multiple links in the camera network, we measure each evaluation metric for all links and average them to get the final topology distance.

			\subsection{Camera Network Topology Training Results}
			\label{subsec:exp:cam_to_cam}
			
			\noindent \textbf{CAM-to-CAM topology inference result}

\begin{figure}[t]
				\centering
				\subfigure[{\scriptsize Connectivity confidences}]{\includegraphics[height=0.35\columnwidth]{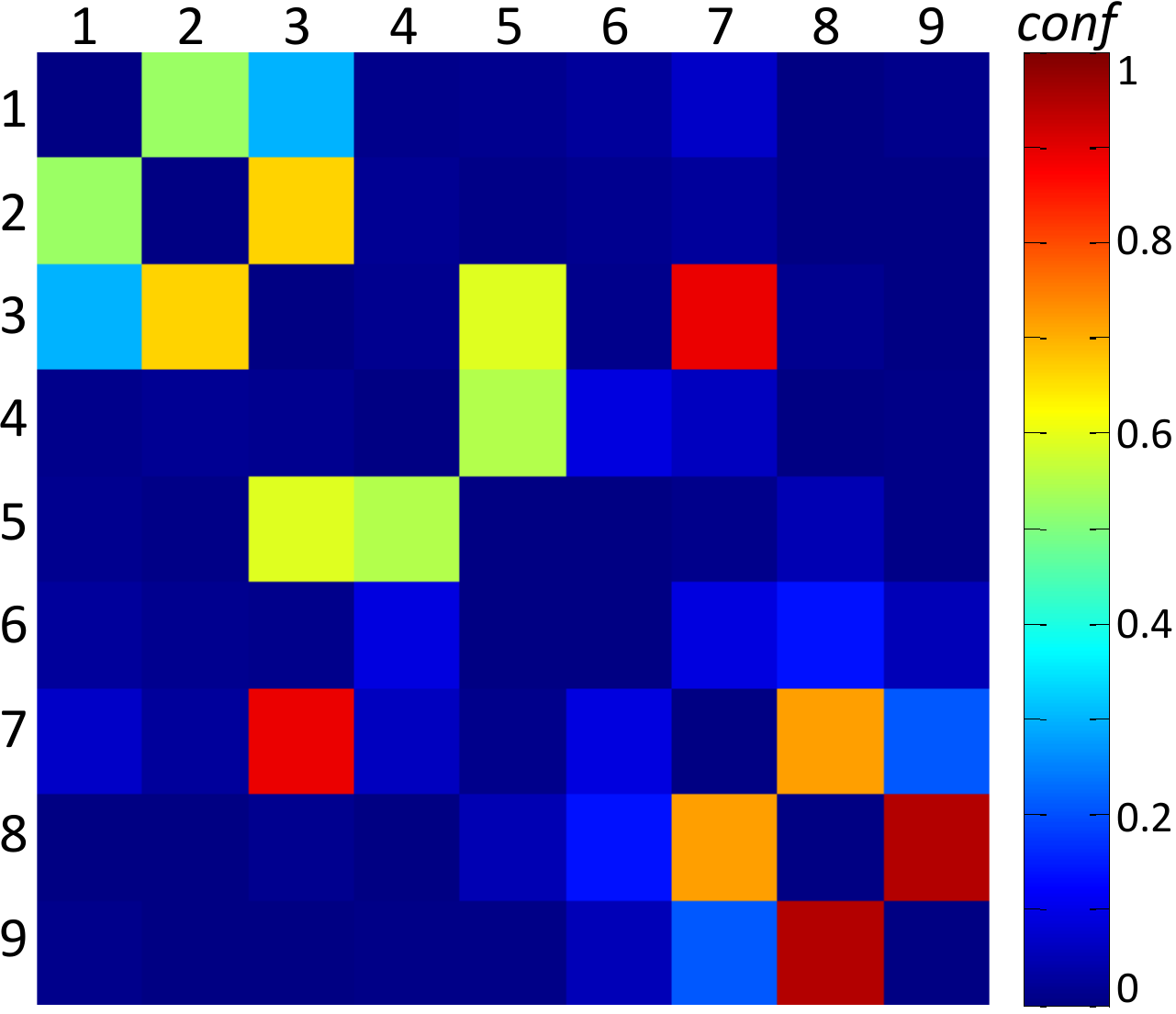}}
				\hspace{20pt}
				\subfigure[Ours]{\includegraphics[height=0.35\columnwidth]{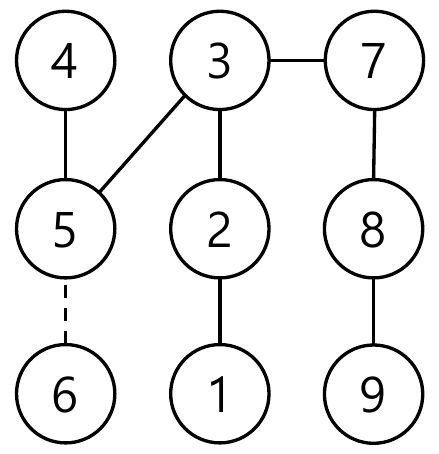}}
				\hspace{3pt}
				\vspace{-0pt}
				\caption{Results of CAM-to-CAM connectivity check. (a) Connectivity confidence map. (b) Valid camera links of two methods (\textit{solid line}: true link, \textit{dotted line}: missing).}
				\vspace{-10pt}
				\label{fig_10}
			\end{figure}

			For all of camera pairs, we illustrate a color map of estimated CAM-to-CAM connectivity confidence in Fig.~\ref{fig_10}~(a). 
			Each row and column indicate the index of the camera. 
			When the confidence value is greater than $\theta_{conf}$, we regard the corresponding camera pair as a valid link. As a result, the valid camera links are drawn as Fig. \ref{fig_10}~(b). Each vertex indicates the index of the camera and valid links are represented by edges. Unfortunately, CAM6 failed to be linked to CAM5. That is because the size of the person image patches is very small due to far distance from the camera; therefore it is hard to distinguish the appearances of people. In addition, CAM6 is quite isolated from other cameras.

			\begin{figure}[t]
				\centering
				\subfigure[]{\includegraphics[height=0.39\columnwidth]{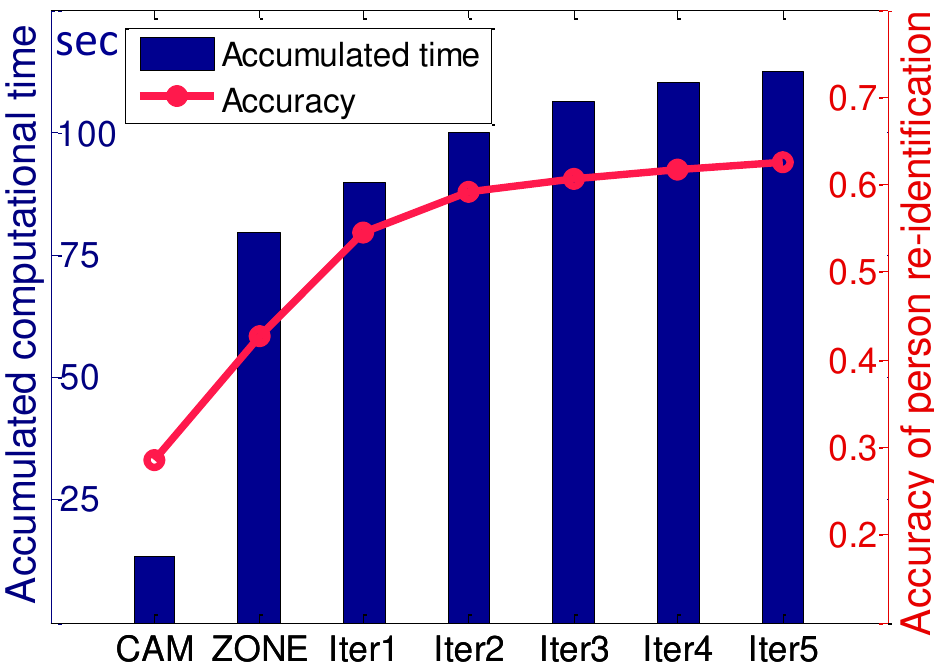}}
				\hspace{3.1pt}
				\subfigure[]{\includegraphics[height=0.39\columnwidth]{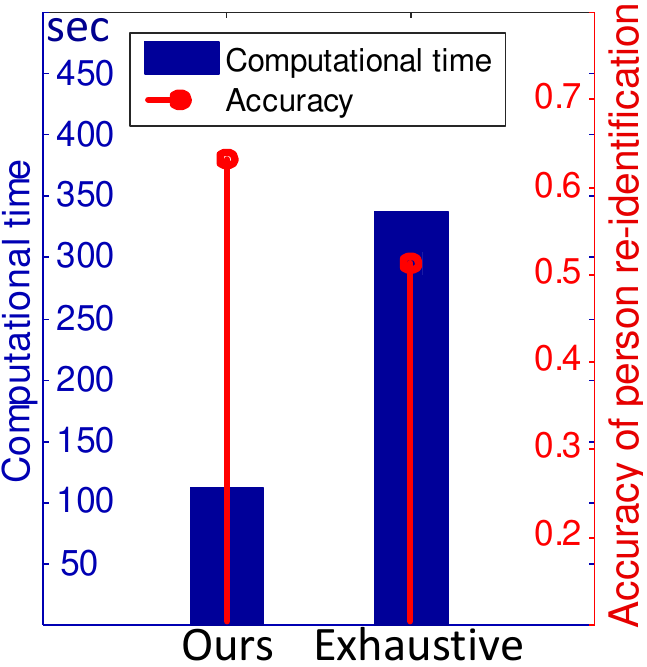}}
				\vspace{-0pt}
				\caption{Results of the person re-identification through the camera topology training (with iteration).}
				\vspace{-0pt}
				\label{fig_11}
			\end{figure}

			\begin{figure*}[t]
				\vspace{-0pt}
				\centering
				\subfigure[Makris~\textit{et al.}~\cite{makris2004bridging}]{\includegraphics[height=0.3\columnwidth]{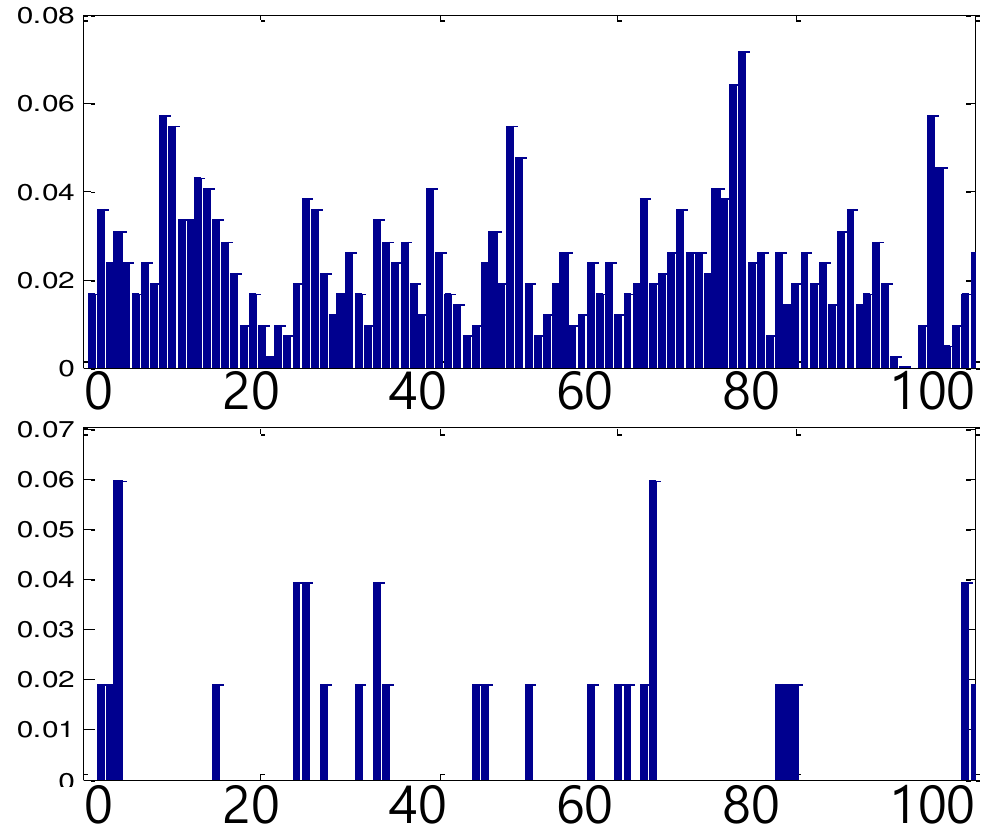}} \hspace{2pt}
				\subfigure[Nui~\textit{et al.}~\cite{niu2006recovering}]{\includegraphics[height=0.3\columnwidth]{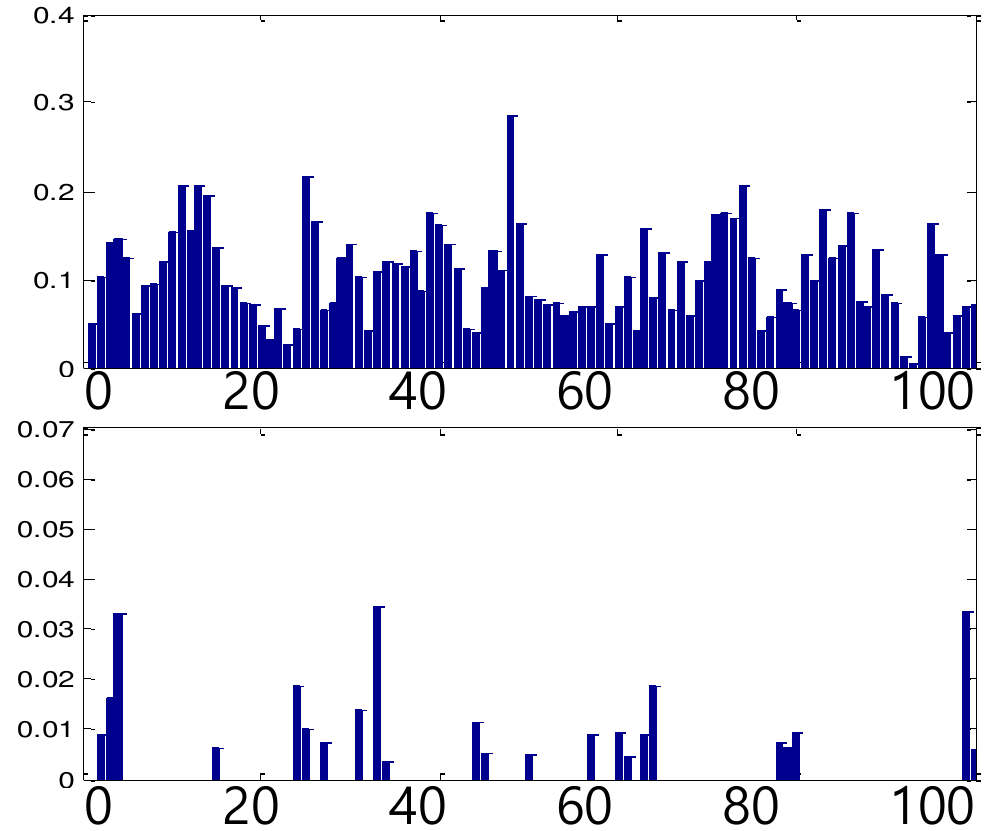}} \hspace{2pt}
				\subfigure[Chen~\textit{et al.}~\cite{chen2014object}]{\includegraphics[height=0.3\columnwidth]{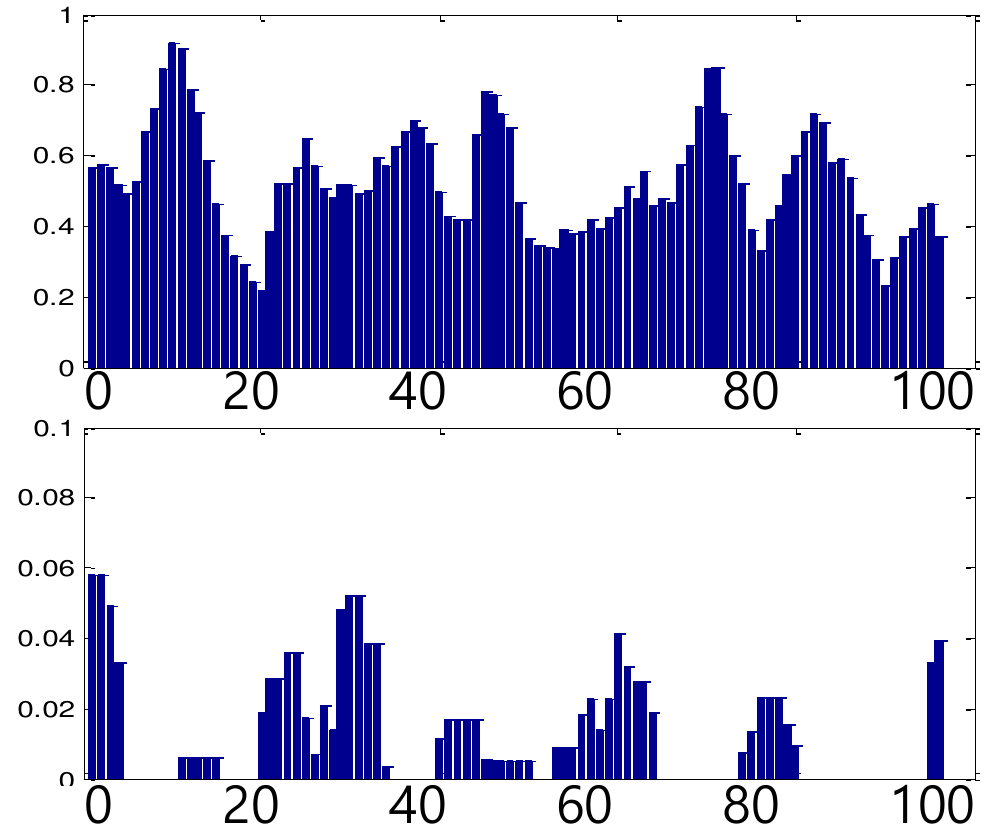}} \hspace{2pt}
				\subfigure[Ours]  {\includegraphics[height=0.3\columnwidth]{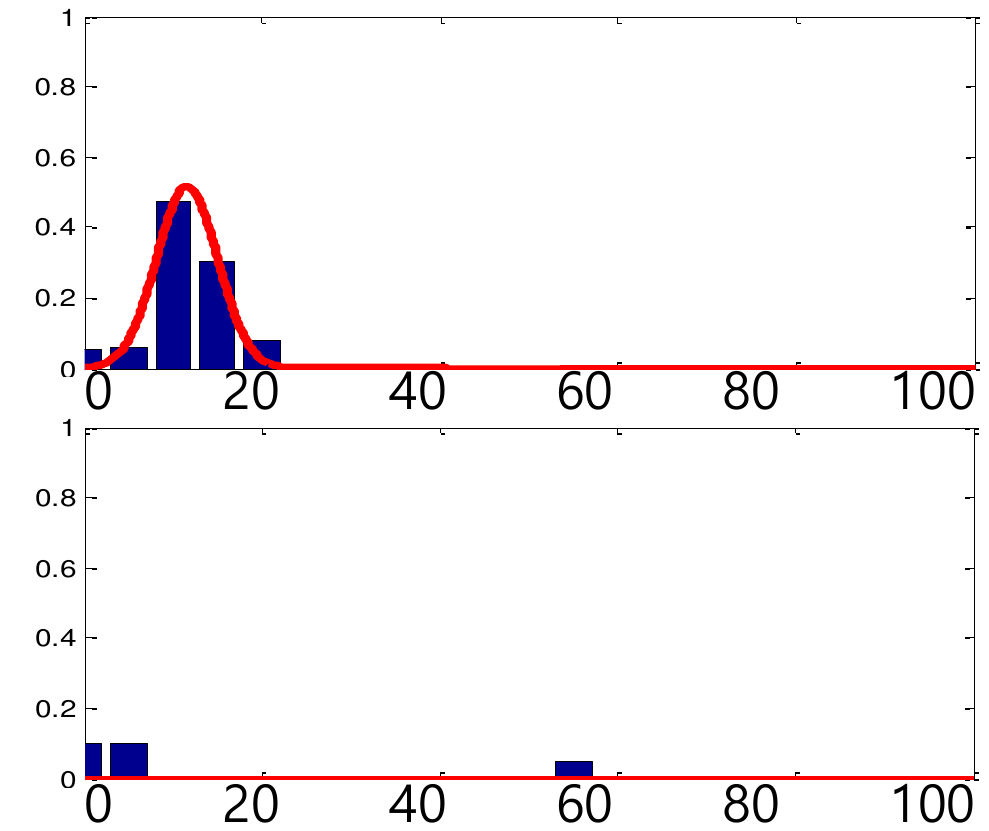}} \hspace{2pt}
				\subfigure[Ground truth]{\includegraphics[height=0.3\columnwidth]{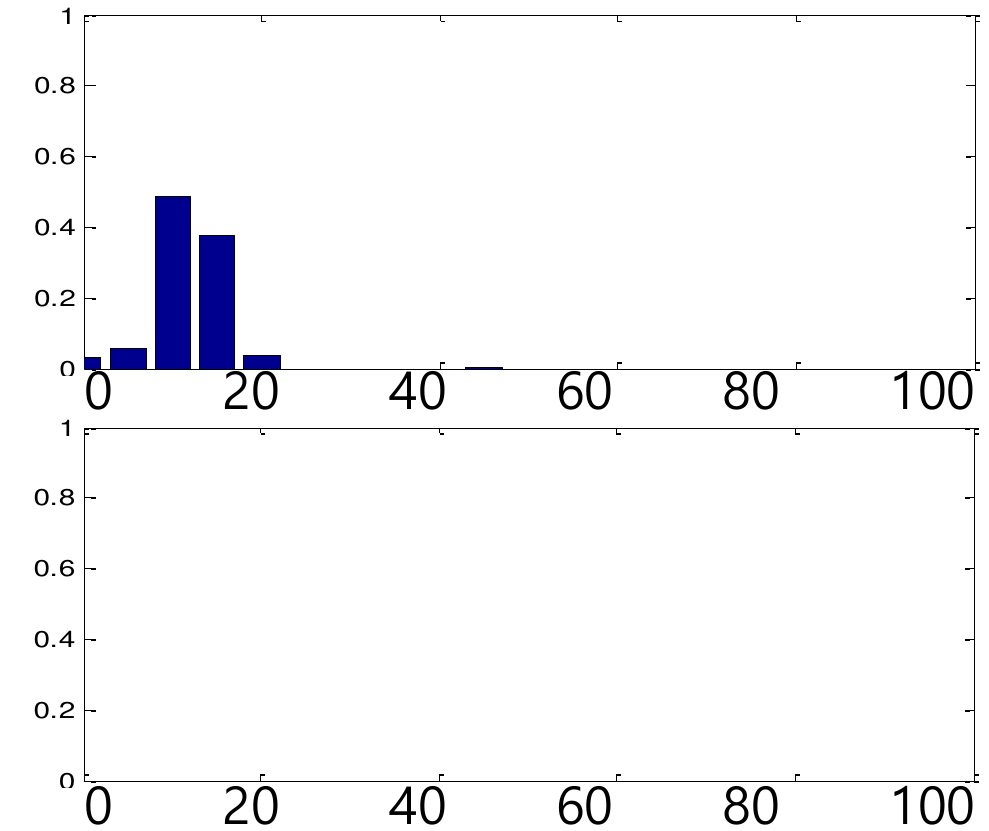}}
				\vspace{-0pt}
				\caption{Comparison of inferred transition distributions and ground truth. First row: Valid link (Exit: \texttt{CAM8,ZONE1}) -- (Entry: \texttt{CAM9,ZONE2}). Second row: Invalid link (Exit: \texttt{CAM3,ZONE1}) -- (Entry: \texttt{CAM7,ZONE2}).}
				\vspace{-0pt}
				\label{fig_12}
			\end{figure*}

			\noindent \textbf{Camera topology training results}
			
			Figure~\ref{fig_11} (a) represents the accuracy of person re-id in each of proposed training steps such as CAM-to-CAM, Zone-to-Zone, and iterative update steps (Sec.~\ref{subsubsec:CAMCAM_con_analy}--\ref{subsubsec:iter_topology_infer}). The accuracy of person re-id is 28.54\% at the beginning, but it is consistently improved by using inferred and refined camera topology information.
			As a result, our method reaches 62.55\% accuracy at the last step of the topology training. 
			In addition, it took only 112.46 seconds to conduct both person re-id and camera topology inference tasks with a large number of people in the nine cameras (using Intel i7 CPU in MATLAB).
			Figure~\ref{fig_11} (b) shows the comparison with a conventional approach which fully compares multiple appearances between people and exhaustively searches the correspondences of people between the entry/exit zones without using camera topology information. 
			It shows lower performance (52.06\% person re-id accuracy) compared to the proposed method, and moreover it takes much more time (337.27 seconds) than ours.

			\begin{table}[t]
				\centering
				\caption{Valid Zone-to-Zone links and ground-truths.} \vspace{-0pt}
				\label{tab_4}
				
				\setlength\tabcolsep{0.75pt}  
				\footnotesize
				\begin{tabular}{c|c|c|c|c|c||c|c|c|c|c|c}
					\noalign{\hrule height 1pt}	
					Exit & Entry                  & $\mu$  & $\mu_{gt}$ &$\sigma$ & $\sigma_{gt}$ & Exit & Entry &                $\mu$ & $\mu_{gt}$ & $\sigma$ & $\sigma_{gt}$\\ \hline \hline
					\texttt{C1,Z1}&\texttt{C2,Z5} & 34.4   &  34.7       & 6.25    & 6.04          & \texttt{C2,Z5}&\texttt{C1,Z1} & 40.4  & 40.4   & 7.62       & 5.93 \\ 
					\texttt{C2,Z2}&\texttt{C3,Z1} & 36.7   &  36.3       & 8.03    & 5.79          & \texttt{C3,Z1}&\texttt{C2,Z2} & 37.6  & 37.0   & 10.3       & 8.90 \\ 
					\texttt{C3,Z2}&\texttt{C5,Z6} & -0.42  &  -0.57      & 3.49    & 3.23          & \texttt{C5,Z6}&\texttt{C3,Z2} & 0.70  & 1.59   & 3.43       & 2.32 \\  
					\texttt{C3,Z3}&\texttt{C7,Z3} & 4.8    &  4.3        & 4.8     & 3.5           & \texttt{C7,Z3}&\texttt{C3,Z3} & 3.75  & 4.68   & 2.16       & 3.04 \\ 
					\texttt{C4,Z4}&\texttt{C5,Z2} & 30.2   &  30.1       & 13.4    & 12.5          & \texttt{C5,Z2}&\texttt{C4,Z4} & 39.5  & 28.6   & 3.82       & 14.8 \\
					\texttt{C7,Z1}&\texttt{C8,Z2} & 28.2   &  28.4       & 21.3    & 6.36          & \texttt{C8,Z2}&\texttt{C7,Z1} & 31.9  & 30.0   & 2.41       & 4.02 \\ 
					\texttt{C8,Z1}&\texttt{C9,Z2} & 11.6   &  11.7       & 4.82    & 4.24          & \texttt{C9,Z2}&\texttt{C8,Z1} & 10.5  & 10.5   & 4.03       & 4.08 \\ 
					\noalign{\hrule height 1pt}
				\end{tabular}
				\vspace{-0pt}
			\end{table}

			A list of valid Zone-to-Zone links inferred by the proposed methods is summarized in Table.~\ref{tab_4}, and the overall results of our method are close to ground-truth $N\left( \mu_{gt},\sigma_{gt}^{2} \right)$.
			The previous methods~\cite{makris2004bridging, niu2006recovering, chen2014object} showed unclear and noisy distributions for both valid and invalid links as shown in Fig.~\ref{fig_12} (a-c).
			On the other hand, our results are very similar with the ground truth (Fig.~\ref{fig_12} (d-e)). 

			
%
%
			
			\subsection{Person Re-identification Test Results}
			\label{subsec:exp:online}
			
			Based on the inferred camera topology in the training stage, we conducted person re-id for the remaining sequence and compared with 
			two different approaches.
			The first approach estimates the camera topology based on \textit{Exhaustive search} in the training stage (11:20 -- 12:20) and
			uses the inferred topology for the re-id test (12:21 -- 13:20). Note that this approach exploits inferred topology but still fully compares multiple appearances of people to find correspondences in the test stage.
			The second approach based on \textit{True matching} estimates the camera topology using ground truth re-id pairs in the training stage and perform the re-id test in the same way with our method. 
			As shown in Table.~\ref{tab_5}, the performance of our re-id test is comparable to that of \textit{True matching}. 
			Our method also outperforms \textit{Exhaustive search} in terms of both re-id and topology accuracies. 
			It supports that our iterative topology update and re-id methods are effective and complement each other.

			\begin{table}[t]
				\centering
				\caption{Performance comparison in training \& test stages.} \vspace{-0pt}
				\small
				\label{tab_5}
				\renewcommand{\arraystretch}{0.95} 
				\begin{tabular}{l|c|c||c}
				\noalign{\hrule height 1pt}
					            & \multicolumn{2}{c||}{Training stage} &  Test stage \\ \cline{2-4}
					            &  Re-id acc   &  Top dist &  Re-id acc  \\ \hline \hline
				Exhaustive	    &  52.1 \%     &  5.620    &  65.6 \%   \\ \hline
				Ours	        &  62.5 \%     &  0.076    &  72.3 \%   \\ \hline \hline
				True matching	&  100 \%      &  0        &  75.6 \%   \\ \noalign{\hrule height 1pt}
				\end{tabular}
				\vspace{-0pt}
			\end{table}

			\section{Conclusions}
			\label{sec:conclusion} 
			
			In this paper, we proposed a unified framework to automatically solve both person re-id and camera network topology inference problems. Besides, in order to validate the performance of person re-id in the practical large-scale surveillance scenarios, we provided a new person re-id dataset called~\texttt{SLP}. We qualitatively and quantitatively evaluated and compared the performance of the proposed framework with state-of-the-art methods. The results show that the proposed framework is promising for both person re-id and camera topology inference and superior to other frameworks in terms of both speed and accuracy.
			
			\section*{Acknowledgement}
			This work was supported by Institute for Information \& communications Technology Promotion(IITP) grant funded by the Korea government(MSIP)(2014-0-00059, Development of Predictive Visual Intelligence Technology).

			{\small
				\bibliographystyle{ieee}
				\bibliography{egbib}
			}

		\end{document}